\documentclass[]{elsarticle}

\usepackage{lineno,hyperref}
\modulolinenumbers[5]

\usepackage{amsmath} %

\usepackage{amssymb} %
\usepackage{booktabs} %

\usepackage{lipsum} %
\usepackage{xcolor}

\usepackage[flushleft]{threeparttable}
\usepackage{array} %

\journal{Pattern Recognition}

\begin{document}

\begin{frontmatter}

\title{Skeleton-based Relational Reasoning \\ for Group Activity Analysis}

\author[ntuaddress]{Mauricio Perez}
\ead{mauricio001@ntu.edu.sg}
\author[sutdaddress]{Jun Liu\corref{mycorrespondingauthor}}
\ead{jun_liu@sutd.edu.sg}
\cortext[mycorrespondingauthor]{Corresponding author}
\author[ntuaddress]{Alex C. Kot}
\ead{eackot@ntu.edu.sg}

\address[ntuaddress]{ROSE Lab, Nanyang Technological University}
\address[sutdaddress]{Singapore University of Technology and Design}

\begin{abstract} %

Research on group activity recognition mostly leans on the standard two-stream approach (RGB and Optical Flow) as their input features.
Few have explored explicit pose information, with none using it directly to reason about the persons interactions. 
In this paper, we leverage the skeleton information to learn the interactions between the individuals straight from it.
With our proposed method GIRN, multiple relationship types are inferred from independent modules, that describe the relations between the body joints pair-by-pair. 
Additionally to the joints relations, we also experiment with the previously unexplored relationship between individuals and relevant objects (e.g. volleyball).
The individuals distinct relations are then merged through an attention mechanism, that gives more importance to those individuals more relevant for distinguishing the group activity.
We evaluate our method in the Volleyball dataset, obtaining competitive results to the state-of-the-art. %
Our experiments demonstrate the potential of skeleton-based approaches for modeling multi-person interactions.

\end{abstract}

\begin{keyword}
Group Activity Recognition, Skeleton Information, Relational Network, Attention Mechanisms.
\end{keyword}

\end{frontmatter}

\section{Introduction} %
\label{sec:intro}

Advances in the field of computer vision and machine learning have naturally led researchers to seek solutions for problems with increasing complexity. Starting from recognition of simple actions performed by a single individual~\cite{Schuldt2004, Blank2005} (e.g., walking, hand-waving, bending) and subsequently moving to mutual-actions being executed by two persons~\cite{Ryoo2009, Yun2012, Kong2012a} (e.g., shaking hands, hugging, punching). Then finally targeting at more advanced activities that encompass many actors at once~\cite{Choi2009, Ibrahim2016, Ramanathan2016}, such as pedestrians queuing or players practicing sports.
Reasoning over scenarios with multiple individuals interacting (or not), can be very useful at many applications, for example surveillance, robot-human interaction and sports analysis.

Research in this field has mainly made use of image visual information -- i.e. RGB and Optical Flow (OF) -- to describe each individual behavior and the interactions among them.
Early works propose methods based on local descriptors and hand-craft features~\cite{Choi2009,Lan2012,Hajimirsadeghi2015}.
Subsequent proposals move to deep-learning based approaches, using Convolutional Neural Networks (CNNs) to extract the low-level features and coupling Recurrent Neural Networks (RNNs) for temporal inference~\cite{Ibrahim2016,Li2017a,Shu2017a,Bagautdinov2017}.
More recent works tackle this problem from a graph-based perspective~\cite{Ibrahim2018,Wu2019a,Tang2019,Lu2020,Yan2020,Yan2020a}.
They use the graph representation to model the group individuals and connections, then apply techniques such as Graph Convolutional Networks (GCNs) to infer the individuals relationships and interactions.
Some of these works do not focus in the interactions at all, while others reason about the interactions at a higher-level in their framework.
Therefore, they lack means to learn the interaction relations at a lower-level in the architecture.

\begin{figure}[!tp]
	\centering
	\includegraphics[width=.9\textwidth]{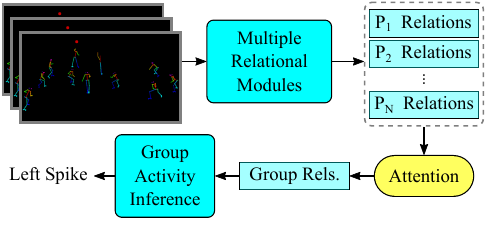}
	\caption{
	Summarized Group Interaction Relational Network (GIRN) pipeline. 
	Skeleton information from all individuals passes through multiple relational modules, each specialized in a different relationship type.
	The inferred relations per individual then pass through our attention mechanism, so they can be merged with different weights into a single descriptor for the group relations.
	Finally, this descriptor is used by our inference module for recognition of the group activity in the scene.
	}
	\label{fig:summary_girn}
\end{figure}

Certain authors aspire to complement or replace the commonly used image visual features with explicit pose information from the individuals~\cite{Azar2018,Lu2019,Chen2020}.
However, these works do not directly leverage the pose information when reasoning about the individuals interactions.
They basically use the pose as an extra/alternative stream for feature extraction. %
Even at the two-person scenario there are not many works that explore human skeleton representation with methods specifically tailored for human interactions~\cite{Yun2012,Ji2014,Wu2018a}. %
Besides, since these proposed methods assume there are only two individuals interacting directly, they lack means to handle the more complex scenario of group activities.
In this scenario there are multiple individuals taking part on a distinguishable collective activity, but performing separate actions which can be similar and related to each other or completely different and independent.

In our work, we propose a method based on skeletons that addresses the aforementioned issues.
Our approach leverages the skeleton information for reasoning over multi-persons interactions. %
A summarized overview of our method is presented in Fig.~\ref{fig:summary_girn}.
We introduce an architecture called Group Interaction Relational Network (GIRN), that operates directly on the joints spatial coordinates, leveraging this representation to infer the interactions among different individuals.
Our architecture, through a pair-wise relational network, can learn distinct relationships relevant for distinguishing the group activity taking place. 
The GIRN's relational modules are guided into specializing for specific types of relationships by consciously choosing which pairs they will be exclusively fed with.
One module focuses on learning the relations between joints from a single individual, while another focuses on the relations of joints from different persons. 
We also try another possible module, which is specialized in the relationship between the human joints and an object of interest (e.g. volleyball).
To further improve the quality of the relations inferred by our modules, we couple our architecture with auxiliary models that allows the knowledge from the individuals actions to be distilled into the network, improving also the group activity recognition.
The GIRN is also equipped with attention mechanisms, which allows our inference modules to attribute higher importance to the relations coming from key individuals.
These individuals have a greater or more distinguishable contribution to the collective activity.

\sloppy{
We validate the effectiveness of our proposed method, for the task of group activity recognition, through comprehensive experiments in the Volleyball dataset~\cite{Ibrahim2016}.
Thus we demonstrate that, when properly leveraged, euclidean space representations can be directly used to reason about the interactions between multiple individuals and also between individuals and objects.
Our method obtains a performance better than several multi-modalities works, and comparable to the state-of-the-art.
}

This paper is organized as follows. In the next section (Section~\ref{sec:related_work}), we present and discuss some of the related works.
In Section~\ref{sec:prop_method} we provide the details to our proposed method: GIRN. 
Experimental details and results are presented in Section~\ref{sec:experiments}, together with insightful discussions.
In Section~\ref{sec:conclusion} we conclude the paper.

\section{Related Work}
\label{sec:related_work}

\subsection{Group Activity Recognition}

Early literature on this topic uses local-descriptors or hand-crafted features to describe the individuals behaviors in space and time.
Choi et al.~\cite{Choi2009} design their own hand-crafted feature based on tracking the individuals with an Extended Kalman Filter, describing their poses with HOG and then constructing a histogram per individual that takes into account its neighbors and their poses.
To capture person-person and group-person interactions, Lan et al.~\cite{Lan2012} model the scene with a structured latent network, on top of a crafted action context descriptor which uses histograms of local-features such as HOG.
Hajimirsadeghi et al.~\cite{Hajimirsadeghi2015} rely on counting the instances of individual actions throughout the clip to define the overall group activity. %

Since the popularization of deep learning and neural networks, research in the field of group activity recognition also explores such techniques.
Ibrahim et al.~\cite{Ibrahim2016} propose an Hierarchical Deep Temporal Model, which consists of two layers of Long Short-Term Memory (LSTM) modules stacked upon each-other.
The base layer contains per-individual LSTMs capturing the person dynamics, and the top layer contains a single LSTM capturing the group dynamics.
The individuals LSTMs have as input CNN features extracted from each player cropped region, throughout sequential frames.
Shu et al.~\cite{Shu2017a} use a similar architecture, but they complement the network with LSTMs for the interactions, and they also propose an energy-based classification layer to replace the softmax layer. 
Bagautdinov et al.~\cite{Bagautdinov2017} proposed method performs the detection and feature extraction at the same time, with a technique based on feeding the frames to a fully-convolutional network which will detect and describe the individuals.
These detections and descriptions are then passed to an RNN for the classification of the group activity in conjunction with the individuals action.
Also somewhat related to detection, Azar et al.~\cite{Azar2019} method is based on generating an intermediate representation that maps directly in the frames the spatial location of the individual actions and the collective activity.
Wu et al.~\cite{Wu2021} propose an optimization scheme to refine the motion information by decoupling the local and global motions, then apply 3D-CNNs for feature extraction.

More recent works keep on using CNN features as input, but focus on inference of the individuals interactions through complex mechanisms over graphs representations of the group.
Ibrahim et al.~\cite{Ibrahim2018} target at learning the players interactions by describing pair-wise relations from matchings defined by hierarchical graphs, then use the descriptions as mid-level representations for each player, before performing the classification with an LSTM at the top.
Wu et al.~\cite{Wu2019a} propose constructing multiple graphs to map the appearance and position relation between players, then use a GCN to extract the players representation used for classification of the individual actions and group activity.
Lu et al.~\cite{Lu2020} build a scene graph by connecting players spatially close to each other, then feed this graph to stacked attention blocks. 
These blocks use graph convolutions to extract person- and group-level interaction representations that will be passed on to Gated Recurrent Units (GRUs) for temporal inference. %

There is a branch of works that tries to go beyond the somewhat standardized input based on CNN features coming from the RGB and optical flow modalities, by incorporating also features coming from pose-based representations.
Azar et al.~\cite{Azar2018} use pose estimation techniques to build a pose heatmap for the scene, consisting on single-channel images where the value of the pixel indicates the presence of any body part. %
The generated image is fed to a CNN for feature extraction.
Lu et al.~\cite{Lu2019} complement their CNN plus GRU pipeline with a spatio-temporal attention mechanism based on skeleton information.
The attention mechanism uses the distance between the poses of the individuals to generate weight coefficients for the deep RGB features from each individual.
Chen and Lai~\cite{Chen2020} also use estimated poses to build a heatmap representation and a CNN for feature learning, but instead of utilizing a single channel for all joints, each joint has its own channel indicating its potential locations.
Gavrilyuk et al.~\cite{Gavrilyuk2020} also apply a pose estimation network for their approach. However, instead of using the estimated pose itself, the output of an intermediary layer is used as additional features for their method. The pose features are fused with RGB and Flow features extracted with the I3D network.
Dasgupta et al.~\cite{Dasgupta2021} employ a similar approach, a pose estimation algorithm is applied to extract features as a mean to infer contextual information, which is added to the appearance information for prediction.

An issue with these approaches is that the pose information is simply regarded as an extra or alternative input feature.
Thus being used as another visual modality that can be fed to an CNN, or through an equivalent feature extractor that feeds on the individual skeleton data.
The useful abstraction of representing the pose as a set of joints spatial coordinates is not directly used for inferring and describing the group interactions.
Different from previous approaches, our proposed method explores such abstraction to learn and describe the relationships between the joints from multiples individuals and their surroundings.
Then our method uses the relationship descriptions to reason about the group activity.

\subsection{Skeleton-Based Interaction Recognition}

Although the literature for action recognition using human skeleton information is quite extensive, it is in fact scarcer when it comes to solutions specifically designed for scenarios with interacting individuals~\cite{Yun2012,Ji2014,Ji2015,Wu2018a,Perez2019a}.
Yun et al.~\cite{Yun2012} propose hand-crafted features, based on pre-defined geometric relations between the joints of two individuals on a series of frames, then feed the computed features to an SVM for classification.
Ji et al.~\cite{Ji2014} group the joints that belong to the same body part to create poselets that describe the interaction of these body parts from both individuals.
Subsequently, these poselets are used to generate a dictionary representation that will be fed to an SVM for classification.
Wu et al.~\cite{Wu2018a} approach this problem using a sparse group lasso on top of features that are based on the individuals joints distance and motion.
Perez et al.~\cite{Perez2019a} avoid using pre-defined relationships by proposing an relational network architecture that can learn directly from the spatial coordinates how the individuals joints relate to each other in a pair-wise manner, then pool these inferred relations to perform the classification.
Zhu et al.~\cite{Zhu2021} target at capturing the interactive features through a dynamic spatio-temporal graph that is different for each frame, and by employing their proposed graph convolution block over it.
Wang et al.~\cite{Wang2021} use estimated poses to generate gray-scale silhouette images that are fed to an CNN for feature extraction. The extracted features are used as input to a model that jointly recognizes the individual actions and the pairwise interactions through an energy function minimization that labels and group the vertices in a complete graph.

Even though some of these works show good performance for recognition of interactions between two individuals, they lack means for handling the complex scenario of group activity. 
This scenario contains multiple individuals with different levels of contribution to the activity itself. 
Our proposed method seeks to bridge the necessary gaps so that the interaction descriptive capability of the skeleton information can also be applied to multi-person settings.

\section{Group Interaction Relational Network}
\label{sec:prop_method}

Our proposed Group Interaction Relational Network (GIRN) is tailored for the recognition of activities involving multiple persons, on which the persons are performing individual actions that can be either similar or distinct to each other.
Inspired by \cite{Perez2019a}, which showed promising results on two-person interaction recognition using skeleton information with a method based on relational networks, we build a novel architecture that can encompass any number of individuals.
This architecture equips important modules necessary to tackle the more complex task of group activity recognition.

In this section, we begin by giving an overview of our architecture and explaining the basic concepts behind it. 
Subsequently we go deeper into the details on how the relational modules operates, and define the different relationship types these modules will be specialized on.
We move on to describe the auxiliary modules, which incorporate the individual actions knowledge into the network.
Finally, we explain the attention mechanisms utilized by our architecture to focus on the relations from the more important individuals.

\subsection{Overview} %

\begin{figure*}[!th]
	\makebox[\textwidth][c]{\includegraphics[width=1.4\textwidth]{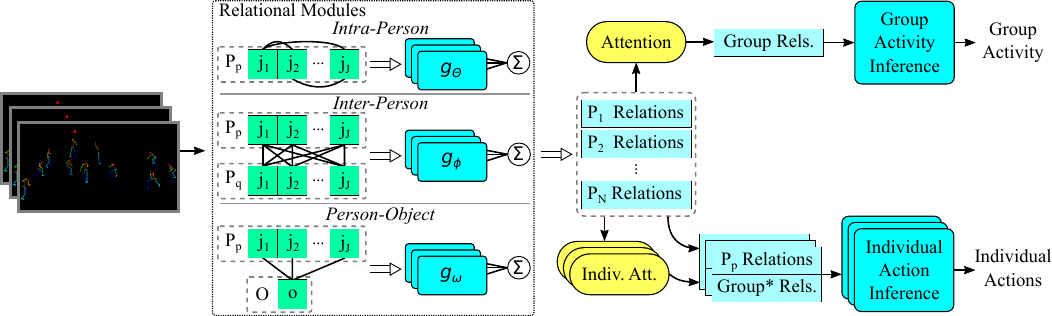}}%
	
	\caption{
		Overview of our proposed method.
		Skeleton information is fed into multiple Relational Modules, each specialized in a different type of relationship. 
		The specialization happens because of the different pairing strategies (illustrated with the joints/objects connections).
		Relations from the same type are averaged pooled ($\overline{\Sigma}$), then concatenated with the others to generate each individual relations.
		These relations are then passed on to our Attention mechanism, which will pool them into a single Group Relations descriptor ($\overline{R}$), at last used by our Group Activity Inference module ($f_G$) to classify the collective activity.
		The auxiliary individual modules runs in parallel to the group branch, and have a similar architecture. 
		The input of the Individual Action Inference module ($f_I$) is the individual relations concatenated to the attention pooled relations from the other individuals ($\overline{\mathcal{R}^{\tilde{p}}}$). 
	}
	\label{fig:detailed_girn}
\end{figure*}

Our method, depicted in Fig. \ref{fig:detailed_girn}, runs directly over the skeleton information, in other words, human joints spatial coordinates. 
Therefore the input features for each person in the scene are the collection of its body joints coordinates, throughout the duration of the activity.
The coordinates are grouped by joint, hence the features of person $p$ is defined as: $P_p=\{j^p_1,j^p_2,\cdots,j^p_J\}$, where $J$ is the total number of joints. Each joint $j^p_i$ is a flat array containing joint $i$ coordinates: 
$j_i=(x_1,y_1,x_2,y_2,\cdots,x_T,y_T)$, where $x_t$ and $y_t$ are the 2D coordinates of the joint $i$ at the frame $t$, and $T$ is the number of frames. Finally, the complete group input can be written as: $G=\{P_1,P_2,\cdots,P_N\}$, for a group with $N$ persons in total.

The joints of each individual is paired in different ways at our relational module, targeting at inferring multiple relationships types for each individual. 
Subsequently we pool the individual relationships as indicated below:
\begin{equation} \label{eq:pool}
	\mathcal{\overline{R}} = \overline{\sum}_{p=1}^{N}\mathcal{R}_p
\end{equation}
where $\overline{R}$ is the final vector with the pooled individual relationships, $\mathcal{R}_p$ is the relationships vectors for person $p$ 
and $\overline{\sum}$ represents any pooling operation (e.g. sum, max, concatenation).
Average pooling is used as the default pooling operation. %

The relationships pooled for the group are then fed to our group activity inference module ($f_G$):
\begin{equation} \label{eq:base_girn}
	GIRN(G) = f_G\left(\mathcal{\overline{R}} \right)
\end{equation}
where in our experiments $f_G$ is implemented as a Multilayer Perceptron (MLP).

The output of $f_G$ is connected to a softmax layer to obtain the group labels predictions, so our architecture can be trained through back-propagation using a cross-entropy loss:
\begin{equation} \label{eq:loss_girn_grp}
	\mathcal{L} = -\frac{1}{N_G} \sum\limits_{g}{\boldsymbol{\hat{y}}_{G,g} \log \boldsymbol{y}_{G,g}}
\end{equation}
where $\boldsymbol{\hat{y}}_{G,g}$ are the ground-truth values for the group activity label, and $\boldsymbol{y}_{G,g}$ are the predictions computed by the softmax layers. $N_G$ is the total number of group activities.

\subsection{Relational Modules} %

The individual relationships used for group activity are extracted with the relational modules described here. These models have the same basic structure $g$, which follows the conceptual idea of relational networks~\cite{Santoro2017}, meaning that each structure has a single pair of objects as input: $g_\theta\left(o_i,o_k\right)$, where $\theta$ represents the parameters set and $(o_i,o_k)$ is the input pair.
For our experiments, $g$ is implemented as a multilayer perceptron (thus $\theta$ represents the learnable weights) and the input is a series of coordinates, such as the joints array $j_i$. 
Moreover, to tailor our relational module to our problem, we explicitly extract the distance and the motion information between the input-pair, and concatenate the extracted information with the coordinates before feeding the input to the first fully-connected layer.%

To better encompass different relationships types relevant to our problem, we use distinct pairing strategies, with every module having its own set of trainable weights: $\Theta$, $\phi$ and $\omega$. This way each module can specialize on identifying and describing a specific type of relationship. The inferred pairs relations are then averaged pooled (represented by $\overline{\sum}$) per relationship type and individual. The strategies adopted (depicted in Fig.~\ref{fig:detailed_girn}) are defined as follows:

\begin{enumerate}[a)]
	\item \textbf{Intra-Person:} Relations of the joints intra-person, i.e. pairs the joints from a single individual with each other.
	\begin{equation}
		\mathcal{R}^{intra}_p = \overline{\sum}_{i=1}^{J}\overline{\sum}_{k=i+1}^{J}{g_\Theta\left(j^p_i,j^p_k\right)}
	\end{equation}
	
	\item \textbf{Inter-Person:} Relations between the inter-person joints, i.e. 
	pairs the joints from a specific individual to the joints from the other individuals.
	\begin{equation} \label{eq:inter-person}
		\mathcal{R}^{inter}_p = \overline{\sum\limits_{q \in \mathbb{C}_p}}\overline{\sum}_{i=1}^{J}\overline{\sum}_{k=1}^{J}{g_\phi\left(j^p_i,j^{q}_k\right)}
	\end{equation}
	where $\mathbb{C}_p$ is the set of individuals connected to the person $p$. Different connection strategies can be applied based on the problem.
	
	\item \textbf{Person-Object:} Relations between a person joints and an object (e.g. volleyball), i.e. pair the joints of an individual to the coordinates of a specific object.
	\begin{equation}
		\mathcal{R}^{object}_p = \overline{\sum}_{i=1}^{J}{g_\omega\left(j^p_i,o\right)}
	\end{equation}
	where the object coordinates are defined analogously as the joints: $o=(x_1,y_1,x_2,y_2,\cdots,x_T,y_T)$.
\end{enumerate}

To use a combination of two or more of the relationships types described above, the relations output vectors are concatenated to generate a new descriptor containing multiple relationships:

\begin{equation}
	\mathcal{R}_p = \left[\mathcal{R}^{intra}_p; \mathcal{R}^{inter}_p; \mathcal{R}^{object}_p \right]
\end{equation}

\subsection{Auxiliary Individual Modules} %

In case the GIRN is being applied on a scenario with well-defined individual actions, where annotations are available during training, our architecture can accommodate the individual label information in order to improve the relationship feature learning.
With some modifications to Eq.~(\ref{eq:base_girn}), the individual module can be defined as:
\begin{equation} \label{eq:base_girn_indiv}
	GIRN_{indiv}^p(G)=f_I\left([\mathcal{R}_p; \overline{\mathcal{R}^{\tilde{p}}}] \right)
\end{equation}
where $f_I$ is implemented as an MLP similar to $f_G$, but with its own set of parameters that are shared between individuals, and $\overline{\mathcal{R}^{\tilde{p}}}$ is the pooled relationships from the individuals other than $p$ (i.e. all except the person $p$). 

To couple the auxiliary individual modules into the training of our architecture, the loss function in Eq.~(\ref{eq:loss_girn_grp}) is redefined as:
\begin{equation} \label{eq:loss_girn_both}
	\mathcal{L} = -\frac{1}{N_G} \sum\limits_{g}{\boldsymbol{\hat{y}}_{G,g} \log \boldsymbol{y}_{G,g}}
	-\frac{2}{N_I \cdot N} \sum\limits_{i}{\boldsymbol{\hat{y}}_{I,i} \log \boldsymbol{y}_{I,i}}
\end{equation}
such that $\boldsymbol{\hat{y}}_{G,g}$ and $\boldsymbol{\hat{y}}_{I,i}$ are the ground-truth values for the group activity and individual action labels, and $\boldsymbol{y}_{G,g}$ and $\boldsymbol{y}_{I,i}$ are the predictions computed by the softmax layers, respectively.
$N_G$ and $N_I$ are the numbers of classes for group activity and individual action respectively.
The loss component coming from the individual actions is divided by the total number of individuals ($N$), so that the group activity recognition be prioritized during the learning process.
However, we multiply this component by a factor of 2 to improve the performance.

\subsection{Attention Mechanisms} %

Although participating on the same group activity, each individual can have different roles during the activity execution itself. Some of these roles can be more important, or simply more discriminating, for determining which activity is in fact happening. 
Therefore, instead of naively average pooling the relations from all the players, an attention mechanism can be used so that more weight can be given to those potentially key individuals.

We opt to equip our network with a dot-product ($\cdot$) based attention~\cite{Luong2015}, hence the attention score equation can be defined as follows:
\begin{equation} %
	a_{G,p} = W_{GQ} \cdot \left( \tanh \left( W_{GK} \mathcal{R}_p \right) \right)
\end{equation}
where $W_{GQ}$ and $W_{GK}$ are trainable weights, respectively responsible for the query and key terms of the dot-product, and $a_{G,p}$ is the group attention score for person $p$.
These scores will be normalized with the softmax function, then be used as weights for a weighted average of the individuals relations being pooled.
The attention pooled relations ($\mathcal{\overline{R}}$) will then be used by the group activity inference module defined in Eq.~(\ref{eq:base_girn}).

Intuitively, an attention mechanism can also be coupled to the auxiliary individual modules. 
A person action should as well be influenced with different degrees by the other persons in the scene. For example it might be a reaction to another individual action, or even be a joint action.
However, since the person importance is relative to a reference person, the individual attention mechanism should not generate a single score for each target person.
It is necessary a more complex mechanism, such as the multi-head attention~\cite{Vaswani2017}, 
which allows computing the targets attention scores considering multiple queries.
Where in our case the query will change according to the reference individual.
To fulfill this requirement, the individual attention mechanism is defined as follows:
\begin{align} %
	key_p &= \tanh \left( W_{IK} \mathcal{R}_p \right) \\ %
	query_p &= W_{IQ} key_p \\ %
	a_{I,p,q} &= query_q \cdot key_p
\end{align}
such that $a_{I,p,q}$ is the individual attention score for person $p$ with reference to individual $q$, and the trainable parameters are $W_{IQ}$ and $W_{IK}$.
Analogously to the group attention, the scores here are softmax normalized and used for weight-average pooling the relations present in Eq.~(\ref{eq:base_girn_indiv}), as a replacement for the average pooling strategy to generate $\overline{\mathcal{R}^{\tilde{p}}}$.

Another component from our method that can benefit from an attention mechanism is the inter-person relationship module. %
When merging the inter-person relations for a particular individual, instead of average pooling its interactions to the other persons with same weight, higher importance can be given to those interactions more meaningful.		
In order to replace this pooling with a weighted sum regulated by an attention mechanism, we define the following equations:
\begin{equation}
	\mathcal{R}^{inter}_{p,q} = \overline{\sum}_{i=1}^{J}\overline{\sum}_{k=1}^{J}{g_\phi\left(j^p_i,j^{q}_k\right)}
\end{equation}
\begin{equation}
	a^{inter}_{p,q} = W^{inter}_{Q} \cdot \left( \tanh \left( W^{inter}_{K} \mathcal{R}^{inter}_{p,q} \right) \right)
\end{equation}
where $\mathcal{R}^{inter}_{p,q}$ are the inter-person relations between individuals $p$ and $q$, $W^{inter}_{Q}$ and $W^{inter}_{K}$ are trainable weights for the query and key terms of the attention mechanism, and $a^{inter}_{p,q}$ is the attention score for the inter-person relations of these individuals.
This way, the individual $p$ overall inter-person relations ($\mathcal{R}^{inter}_p$) can be computed as a weighted sum of the inter-person relations between $p$ and the other individuals ($\mathcal{R}^{inter}_{p,q}$), given the attention score $a^{inter}_{p,q}$.
\section{Experiments}
\label{sec:experiments}

\subsection{Dataset}

To evaluate our method, we use the Volleyball dataset~\cite{Ibrahim2016}, which consists of 4,830 snippets (41 frames each) of group activities from professional athletes playing Volleyball. 
The dataset annotation comprises labels referring to four different activities (set, spike, pass and winpoint), in conjunction with an indication on which team is performing the activity (left or right), totaling eight distinct labels (e.g. right spike, left winpoint).
The dataset authors also annotated the individuals actions (nine in total): waiting, setting, digging, falling, spiking, blocking, jumping, moving and standing. 
Together with the players action labels, they also provide the bounding box locations at the central frame of the snippet. %
For evaluation, the authors have separated the data into train, validation and test splits, and report the group activity classification accuracy at the test split.
We follow the same protocol as the authors on our experiments.

\begin{figure}[ht!]
	\centering
	\setlength{\tabcolsep}{0.1em}
	\begin{tabular}{cc}
		\includegraphics[width=.48\textwidth]{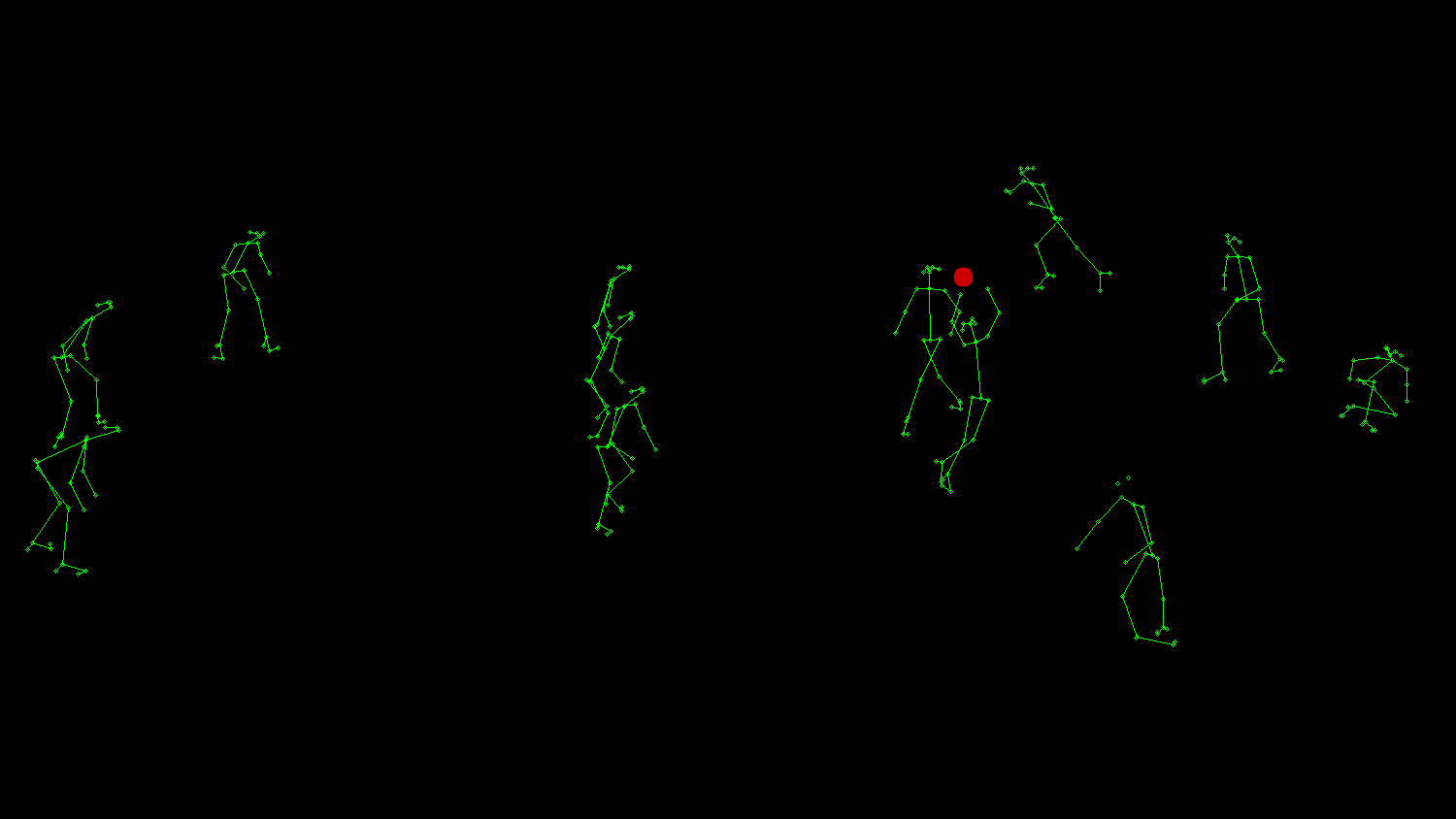} & 
		\includegraphics[width=.48\textwidth]{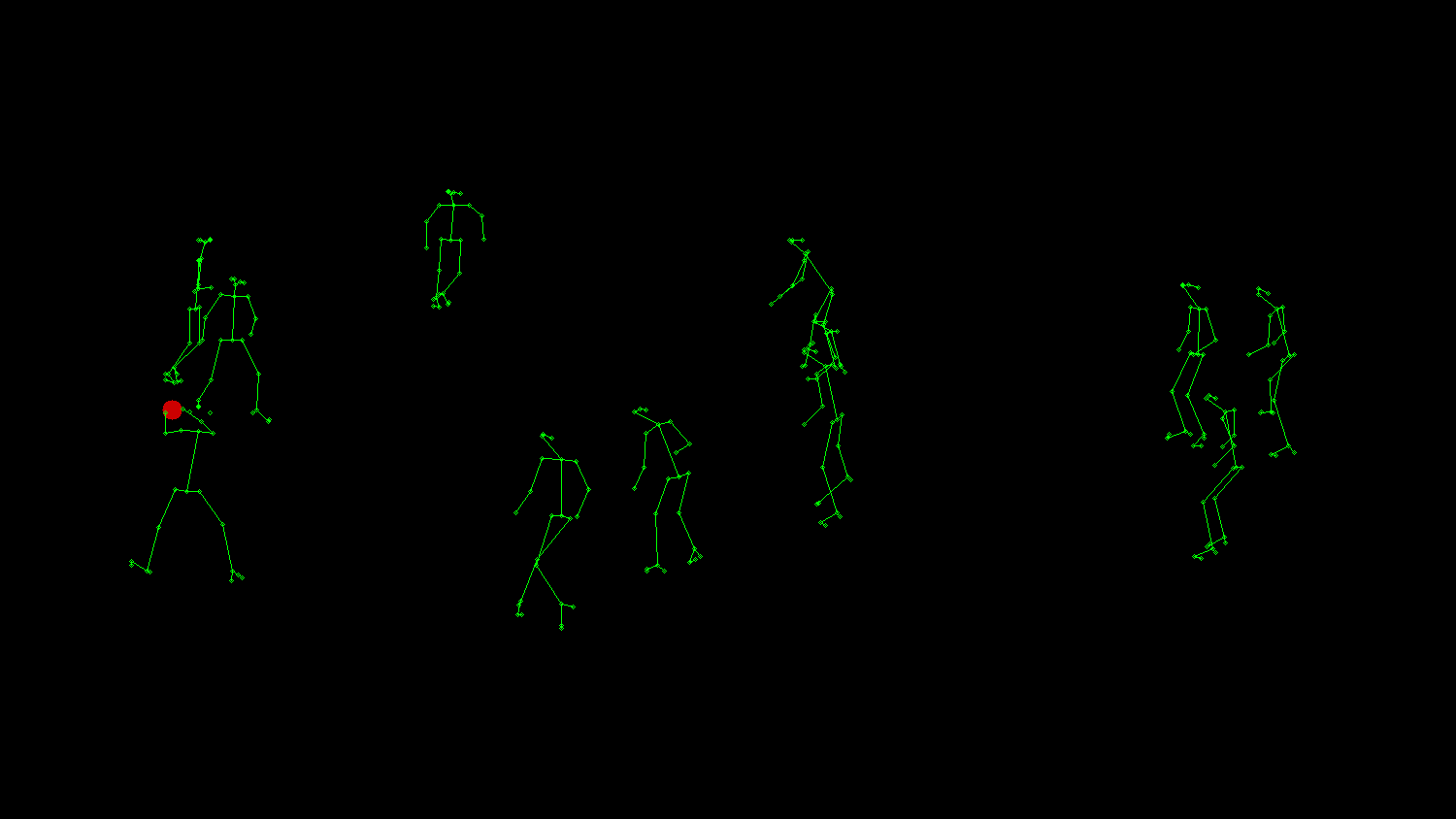} \\
		(a) Right Set &  (b) Left Pass \\[1.em]
		
		\includegraphics[width=.48\textwidth]{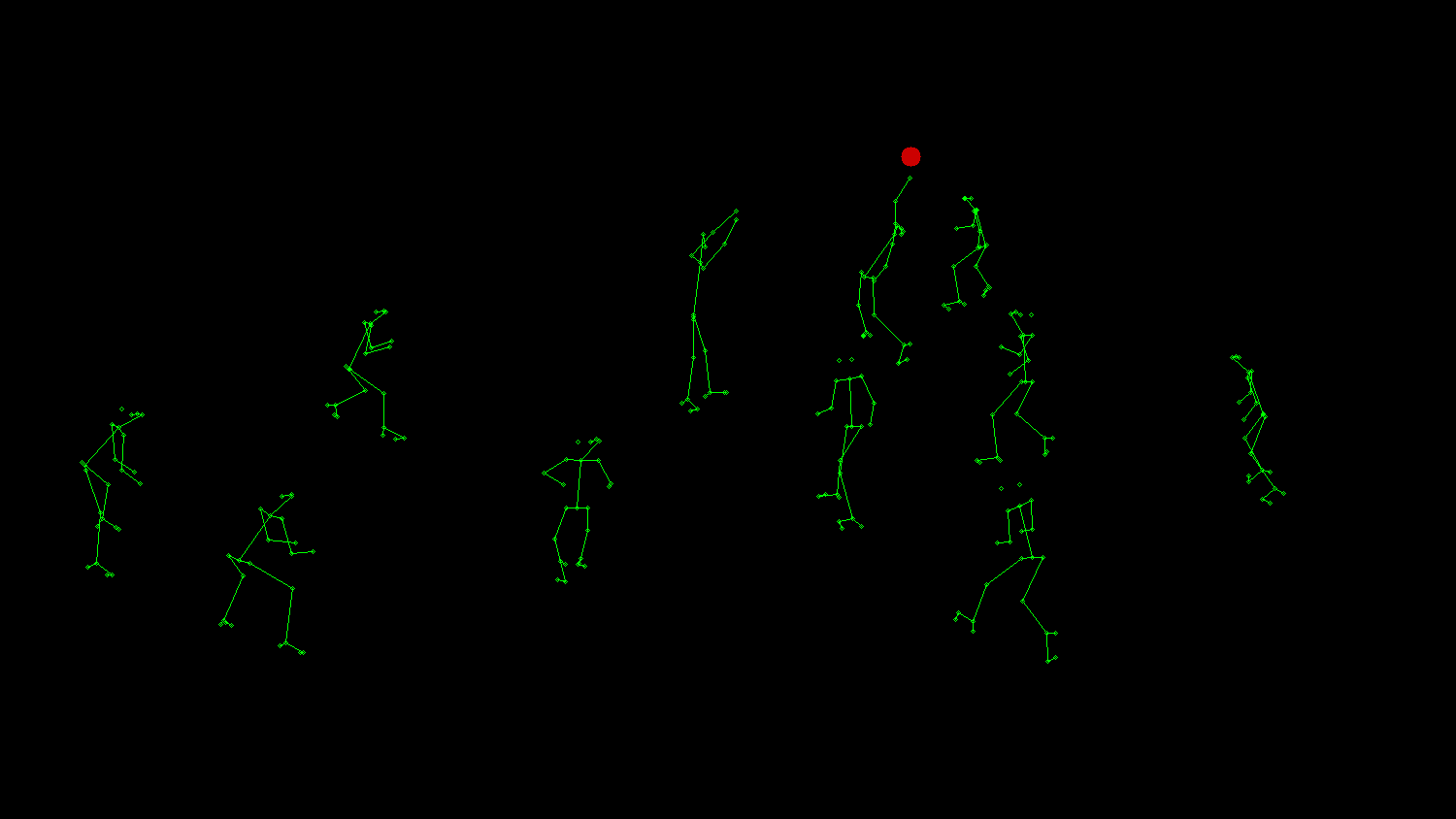} & 
		\includegraphics[width=.48\textwidth]{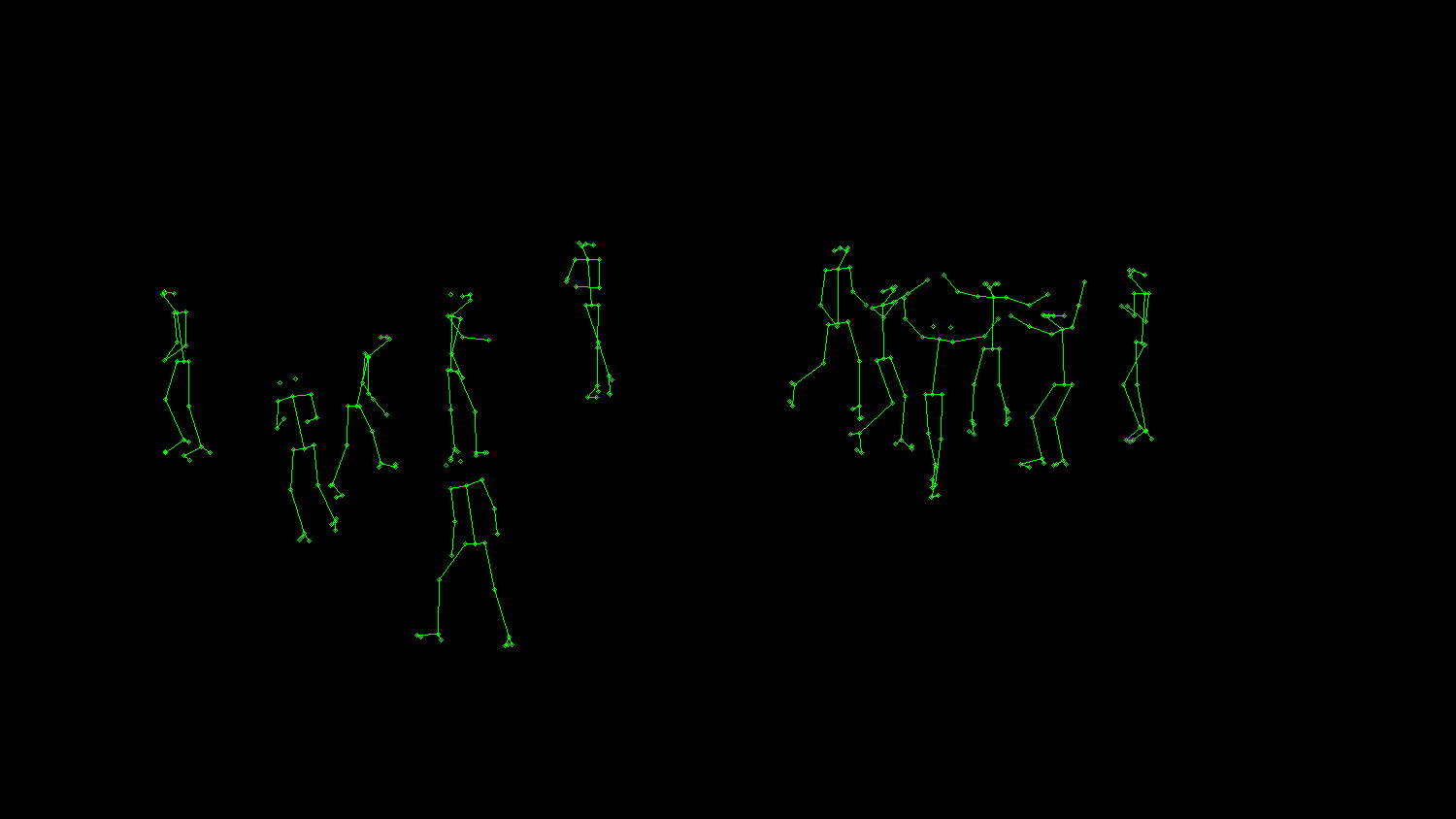} \\
		(c) Right Spike &  (d) Right Winpoint \\[1.em]
	\end{tabular}
	\caption{Examples of estimated poses and annotated ball coordinates from Volleyball dataset. Contains examples from four distinct group activities: (a) Right Set;  (b) Left Pass; (c) Right Spike; and (d) Right Winpoint.}
	\label{fig:volley_skels_examples}
\end{figure}

Since our proposed method is skeleton-based, we use a pose estimation method to extract the skeleton information from the Volleyball dataset.
Examples of extracted poses are presented in Figure~\ref{fig:volley_skels_examples}.
As it can be seen, the distinctive poses from the players provide highly informative data for inferring which actions are the individuals performing and, consequently, what is the group activity happening in the scene.
For example: in (a), the player setting has an unique pose with both arms raised and jumping towards the ball, meanwhile some of its teammates are running towards the front in preparation to spike the ball; 
in (b), the player passing is also reaching for the ball with both arms, but from a lower height (the ball is below its head), and with the hands closer to each other;
in (c), the player spiking is jumping with a single arm moving towards the ball, meanwhile the opponent have its arms raised for blocking the incoming spike; 
in (d), the players from the team with the winpoint are celebrating by gathering with their arms raised in an open angle.

\subsection{Implementation Details}
\label{subsec:impl_details}

For our experiments, the skeleton information is extracted using OpenPose~\cite{Cao2018}, with the \textit{BODY\_25} model and net resolution of \textit{1312$\times$736}.
We apply basic ad-hoc heuristics to attribute the extracted poses to each individual, and impute short-gap missing detections with interpolation.
The skeletons are then normalized by removing the estimated camera motion between frames, and changing the origin to the coordinate central to all poses. %
For Intra-Person and Person-Object relationship types we use the following seven joints: Nose, Neck, Mid Hip, Left/Right Wrists and Left/Right Ankles. For Inter-Person we restrict to: Left/Right Wrists and Left/Right Ankles.
We select this joints subsampling through preliminary experiments.
The poses are randomly mirrored during training (by inverting the x-coordinates).
This works as a data-augmentation technique to reduce the impact of over fitting. 
The group label is also inverted (left/right) when the skeleton is mirrored.

Although the poses are estimated for all 41 frames, not all of them are used when building the input for our method. 
We alternately sample the frames, since that leads to better performance for our models.
In other words, the joints' arrays $j_i$ are constructed by using the coordinates from 21 of the 41 frames available, by picking one and skipping the next.

Unfortunately, for the estimation of the ball coordinates there is no reliable tracker yet, to the best of our knowledge. 
Current research in ball tracking still faces difficulties generalizing to in-the-wild scenarios, where there is no information about the camera or any type of calibration~\cite{Kamble2019}.
To overcome this issue, and be able to assess the potential of our Person-Object relationship type, we use manually annotated ball coordinates\footnote{Ball annotation: \url{https://drive.google.com/file/d/1urZpZiiepC85JD1u3VeURgUpztRgI0yl}}.

As it was mentioned previously, our relational and inference modules are implemented as MLPs. The relational module ($g$) consists of four fully-connected layers, first three with 1000 units and the fourth with 500. For the inference ($f$) module, we use three fully-connected layers with 500, 250 and 250 units respectively. 
The group and individual attention mechanisms parameters ($W_{GQ}$, $W_{GK}$, $W_{IQ}$ and $W_{IK}$) have the same dimensions as the input relations vectors ($R_{p}$).
For the inter-person relations attention, the parameters ($W^{inter}_{Q}$ and $W^{inter}_{K}$) have the same dimensions as the inter-person relations vectors ($R^{inter}_{p,q}$).
Our whole network is trained through back-propagation with Adam optimizer and learning rate value of 1e-4. Weight initialization coming from a truncated normal distribution with 0 mean and 0.045 standard deviation.
We use a dropout rate of 0.25 for the $f$ module and the attention parameters, no dropout at $g$.
Additionally to the individual loss weighting factor defined in Eq. (\ref{eq:loss_girn_both}), we also apply weights specific to class, since the number of instances for each of the individual actions is highly imbalanced (Standing represents almost 70\%, Jumping only 0.6\%).
The relationship types are first trained separately, then their specific set of $g$ parameters ($\Theta$, $\phi$ and $\omega$) is fine-tuned while the upper layers (attention mechanisms and $f$ modules) are trained from scratch with the concatenated relations.

The individuals connection strategy used by the inter-person relationship type, responsible for defining $\mathbb{C}_p$ in Eq.~(\ref{eq:inter-person}), is tailored for the players structure in volleyball matches.
Individuals are first split into two teams, then into two sub-groups: front and back.
The \textit{front} sub-group consists of the three players positioned closer to the volleyball net and the \textit{back} sub-group are the remaining three, positioned furthest from the net.
The players at the \textit{back} are only connected to those on the same team, meanwhile the players in the \textit{front} are also connected to the opponent players positioned in the other team's \textit{front}.
In case there are players missing, we leave unfilled spots in the \textit{back} sub-groups.
Connecting all the players would be counter-effective because the features from the important interactions could vanish due to the average operation with so many other interactions, possibly less relevant. Moreover, the number of pairs grows geometrically with the number of individuals connections and the number of joints used, becoming computationally prohibitive at some cases.
In Subsection~\ref{subsec:inter_connect} we provide empirical results addressing these issues, that further support the connection strategy adopted.

\subsection{Results}

\subsubsection{Number of Frames for Poses Sampling}

\begin{figure}[!ht]
	\centering
	\includegraphics[width=.7\textwidth]{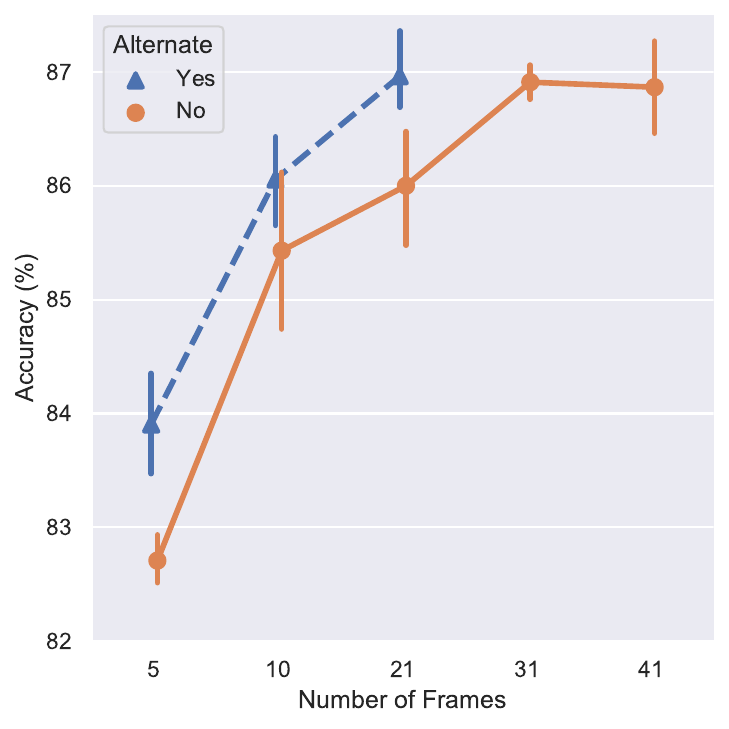}
	\caption{
	Evaluation of our approach based on the number of frames used for sampling the poses.
	Results using the baseline model with Intra-Person relationship type. 
	The ``Alternate'' curve indicates whether the frames are alternately sampled, in contrast to be sampled sequentially without skipping any. The frames are always sampled around the middle frame.}
	\label{fig:num_frames}
\end{figure}

We start with experiments regarding the impact in performance by varying the number of frames from which the poses are sampled from.
Since our approach is skeleton-based, using pose information from more frames has little impact in the input data size and computational requirements for our method.
Different from visual information (e.g. RGB and Optical Flow), employed by most of the previous works.
Therefore, we conduct the study on number of frames described here.
For this experiment we use our baseline model (without individual modules and attention mechanisms) with the Intra-person relationship type.
The poses are always sampled from the frames around the middle frame.
In contrast to sequentially sample from all frames, the poses can be sampled alternately. 
Which means poses sampled by alternating between using a frame and skipping the next.
The number of frames for alternately sampling is limited to 21 due to the total number of frames available, which 41 for the Volleyball dataset.
The results are shown in Figure~\ref{fig:num_frames}.

Without alternate frame sampling, increasing the number of frames used also increases the accuracy obtained.
However, increasing from 31 frames to 41 leads to overfitting and therefore a lower performance in average.
Interestingly, if the poses are sampled from alternate frames, a higher performance can be obtained even though using less frames.
This occurs because alternately sampling the frames allows our approach to extend the temporal range, but with higher diversity between consecutive poses, which helps reducing overfit.

\subsubsection{Ablation Experiments}

Here we conduct experiments
to evaluate the efficiency of the relationship types independently, and their positive complementarity when fused. In these experiments, we also incrementally assess the impact of the different parts of our architecture.
The results are shown in Table~\ref{tab:ablation}.

\begin{table}[ht]
	\centering
	\begin{threeparttable}
	\caption{Ablation experiments on Volleyball dataset. Accuracy performance of our proposed GIRN with different relationships types and cumulative addition of components.}
	\begin{tabular}{lccc}
		\toprule
		\textbf{Relationships} & \textbf{Baseline} & \textbf{+ Indiv. Modules} & \textbf{+ Attention} \\ \midrule
		Intra-Person 				& 86.7\% & 87.6\% & 88.0\% \\
		Inter-Person 				& 83.5\% & 84.4\% & 84.7\% \\
		Intra+Inter  				& 87.2\% & 88.0\% & 88.4\% \\
		\midrule
		Person-Object				& 85.5\% & 87.1\% & 87.6\% \\
		Intra+Inter+Object	& 90.1\% & 91.8\% & 92.2\% \\
		\bottomrule
	\end{tabular}
	\label{tab:ablation}
	\end{threeparttable}
\end{table}

The baseline for our proposed method makes use only of the group activity labels during training, and the individual relations are average pooled with equal weights before being fed to the group activity inference module.
Among the pose-only relationships, Intra-Person obtains a notable better performance than Inter-Person. 
However, their complementarity consistently leads to an improvement when fused (Intra+Inter).
The quality of these relations are meaningfully enhanced when the auxiliary individual modules are added to the baseline, allowing the group activity module to obtain an increase of almost 1 percentage point of accuracy.
Subsequently, replacing the naive average pooling with our defined attention mechanisms further increase the performance, with 88.4\% of accuracy being the highest result from our GIRN method when using only the pose information. 

Aggregating the ball location information to our method allows the Person-Object relationship type to be also used. 
The inferred relations coming from this module show promising discriminating properties, obtaining a performance close to the Intra-Person relationship.
Moreover, its high complementarity to the previous two relationship types leads to a significant boost in performance, with 92.2\% of group activity recognition accuracy.
Hereinafter we denote our method with the complete architecture by $\text{GIRN}_{intra+inter}$, for the relations using only the estimated poses, and $\text{GIRN}_{intra+inter+obj}$, when including the relations using the annotated ball information.

\subsubsection{Inter-Person Connectivity and Number of Joints} %
\label{subsec:inter_connect}

The performance for the inter-person relationship type is dependent on the predefined connectivity among the individuals and the number of joints selected.
The results for the ablation experiments above follow the configuration detailed in  Subsection~\ref{subsec:impl_details}, with a subsampling of four joints and a densely connection strategy (although not fully-connected).
Here we explore the impact on the performance of different connection strategies and joints subsampling in conjunction.
We also evaluate the inter-person attention mechanism.

For the joints subsampling, we range from two to seven joints, starting from the upper limbs (Wrists) and incrementally adding joints from the spine (Neck, MidHip and Nose) and lower limbs (Ankles).
The connection strategies are based in the player position in the volleyball court.
In a volleyball game, there are six players per team, arranged facing the net in two rows and three columns.
For simplicity, we refer to the players positions based if they are in the front or back row, and if they are in the middle column or in the exterior columns.
The explored connection strategies are: %
\begin{itemize}
    \item \textbf{Full Connectivity}: Every player is connected to all other players.
    \item \textbf{Dense Connectivity}: All players are connected to their teammates, and players in the front row are also connected to the opponent team front row players.
    \item \textbf{Moderate Connectivity}: Similar to Dense, but players at the exterior columns
    have less connections (they are not connected to the players in the opposite exterior column).
    \item \textbf{Sparse Connectivity}: Similar to Moderate, but players at the middle column also have less connections (they are not connected to the players at their diagonals, i.e. the players in the exterior columns that are not on the same row as the player).
\end{itemize}

\begin{table}[t!]
    \centering
    \begin{threeparttable}
    \caption{Evaluation of Inter-Person relationship type with different players connectivity degrees and varying number of joints per subsampling.}
    \begin{tabular}{lcccccc}
        \toprule
    Connectivity    & 2-joints & 3-joints & 4-joints & 5-joints & 6-joints & 7-joints \\
        \midrule
    Full            & 83.4\%   & 83.8\%   & 82.9\%   & -        & -        & -        \\
    Dense           & 84.1\%   & 84.4\%   & \textbf{84.7\%}   & 83.7\%   & 83.8\%   & -        \\
    Moderate        & 83.1\%   & 83.4\%   & 82.9\%   & 83.3\%   & 83.3\%   & 82.7\%   \\
    Sparse          & 81.9\%   & 82.3\%   & 82.9\%   & 82.1\%   & 81.9\%   & 83.5\%   \\
	\midrule
	Full-Att        & 84.5\%   & 83.8\%   & \textbf{85.5\%}   & -        & - & - \\
	Dense-Att       & 84.4\%   & 84.7\%   & 85.2\%   & 84.3\%   & 83.9\%   & -        \\
        \bottomrule
    \end{tabular}
    \begin{tablenotes}
        \scriptsize
        \item \underline{2-joints}: Left/Right Wrists; \underline{3-joints}: 2-joints + Neck; \underline{4-joints}: 2-joints + Left/Right Ankles;
        \item \underline{5-joints}: 4-joints + Neck; \underline{6-joints}: 5-joints + MidHip; \underline{7-joints}: 6-joints + Nose.
    \end{tablenotes}
    \label{tab:inter_joints}
    \end{threeparttable}
\end{table}

The results are reported in Table~\ref{tab:inter_joints}. 
For some combinations of connectivity and joints subsampling there are so many input pairs that it becomes computationally infeasible to run.
Thus there are absent values in the higher connectivity and joints cells (e.g. Full Connectivity with 5-joints, Dense Connectivity with 7-joints).
Nonetheless, it can be seen that the performance does not always improve with the increase of connectivity or the number of joints. %
For example, using 5 or more joints seems to be detrimental to the performance with Dense connectivity.
Moreover, the optimal number of joints depends on the level of connectivity. %
Four joints seems to be ideal for Dense, but not ideal for the other connectivity levels.
In general there seems to be a reverse performance correlation between the degree of connectivity and the number of joints, i.e. the optimal number of joints increases as the degree of connectivity decreases.
Overall, among the results without attention, the dense connectivity degree obtains the highest accuracy results, with 4-joints being its optimal subsampling.

Employing the attention mechanism for merging the inter-person relations leads to an improvement in accuracy for most cases, both with the Full and Dense connectivity.
We only report the results for the Full and Dense connectivity, since their version without attention is equivalent or superior to the other connection strategies.
The Full connectivity strategy greatly benefits from the inter-person attention mechanism, achieving the highest performance for the inter-person relationship, with 85.5\% of accuracy when using 4-joints.
However, extracting inter-person relations between all the individuals is computationally expensive, particularly in scenarios with many people involved.
By consciously selecting the connectivity between the individuals, the specialized connection strategy can reduce the computation cost with little sacrifice to performance.
As it can be seen, densely connecting the individuals also shows an improvement by employing the attention mechanism, obtaining a performance close to the Full connectivity.
This is possible in group activity scenarios where the individuals interact following a certain type of structure, as in the Volleyball scenario.
Therefore, through our method, such connection strategies present themselves as interesting possibilities to obtain a good performance while reducing the computational cost.
In other scenarios, without a well-defined interaction structure but with less individuals involved, the Full connectivity with inter-person attention is a practical and effective alternative.

Since we are evaluating our method in the Volleyball dataset, when fusing the different relationship types we use the Dense connectivity strategy without inter-person attention.
As the Dense connectivity strategy provides a good performance at a lower computational cost.

\subsubsection{Robustness Evaluation} %
\label{subsec:robust_eval}

To analyze the impact on performance of using inaccurate ball coordinates, as if they were coming from ball-tracking algorithms, we conduct experiments adding different types of noise to this information and evaluating our complete method over it: Intra+Inter+Object relationships with individual modules and attention mechanisms.
This investigation is important to assess how robust our method is with regard to the ball information, i.e. how precise this information has to be such that GIRN leverages it when performing group activity recognition.
The results from these experiments are presented at Table~\ref{tab:pixel_noise} and Table~\ref{tab:dropout_noise}. %
Further details and discussion are provided subsequently.

\begin{table}[ht]
    \centering
    \begin{threeparttable}
    \caption{Evaluation of robustness against misplaced ball localization. 
    Zero-mean Gaussian generated noise is added to the ball coordinates, with variable standard deviation values indicated by the displacement noise in terms of pixels.
    }
    \begin{tabular}{lccccccc}
        \toprule
        \textbf{Noise Std Dev} %
            & 0\textit{px}    & 20\textit{px}   & 40\textit{px}   & 60\textit{px}   & 80\textit{px}   & 100\textit{px}  & 120\textit{px}  \\% & 140    & 160 
        \midrule
        \textbf{Accuracy\tnote{$\ast$}} %
            & 92.2\% & 92.0\% & 91.4\% & 90.8\% & 89.3\% & 89.1\% & 87.6\% \\% & 86.3\% & 85.1\% \\
        \bottomrule
    \end{tabular}
    \begin{tablenotes}
        \scriptsize
        \item[$\ast$] Accuracy averaged over five repetitions per parameter.
    \end{tablenotes}
    \label{tab:pixel_noise}
    \end{threeparttable}
\end{table}

\begin{table}[ht]
    \centering
    \begin{threeparttable}
    \caption{Evaluation of robustness against ball miss-detections. %
    Ball coordinates at each frame is randomly set to zero according to the dropout chances indicated.
    }
    \begin{tabular}{lccccccc}
        \toprule
        \textbf{Dropout Chance} & 0\%    & 5\%    & 10\%   & 15\%   & 20\%   & 25\%   \\
        \midrule
        \textbf{Accuracy\tnote{$\ast$}} & 92.2\% & 91.2\% & 90.3\% & 89.6\% & 88.8\% & 87.7\% \\
        \bottomrule
    \end{tabular}
    \begin{tablenotes}
        \scriptsize
        \item[$\ast$] Accuracy averaged over five repetitions per parameter.
    \end{tablenotes}
    \label{tab:dropout_noise}
    \end{threeparttable}
\end{table}

We start by studying the impact of misplaced localization, through addition of Gaussian noise to the coordinates. The generated noise has zero mean and variable standard deviation parameters, where the standard deviation values represent the displacement in terms of pixels from the ball actual coordinates.
The majority of the videos have a resolution of 1280$\times$720, but a few have 1920$\times$1080. %
To keep the added displacement proportional, we multiply the indicated pixel value by $1.5$ times when generating the noise for the latter case.
As it can be seen at Table~\ref{tab:pixel_noise}, our method is considerably robust to inaccurate ball coordinates, it can still obtain more than 90\% of accuracy even with a displacement noise of standard deviation set to 60 pixels (approximately $1/20th$ or $1/12th$ of the frame width and height respectively)s.
The use of the ball information stops being advantageous (same performance as Intra+Inter) only when the standard deviation value reaches 120 pixels, which is a very high displacement noise (approximately $1/10th \times 1/6th$ of the frame resolution).

Our next experiment, reported at Table~\ref{tab:dropout_noise}, targets at measuring the impact of ball missed detection at some frames, therefore having an input with coordinates equal to zero at different timesteps. 
To accomplish this, we randomly set the ball coordinates to zero according to distinct dropout chances and feed it to our method for testing.
Our method is more sensitive to this type of noise, however it remains capable of achieving more than 90\% accuracy regardless of 10\% of the coordinates being set to zero.
In fact, a quarter of the sequence has to be missing for our method to stop obtaining a performance better than using only the skeletons.
Moreover, taking into consideration that the previous experiment validates the robustness of our method to inaccurate coordinates, to reduce the negative impact to performance the miss-detections can be imputed with interpolated values from the acquired detections.

In conclusion, the results from these experiments indicate that our method is sufficiently robust to noise in the ball coordinates, not requiring a precise ball detection in all frames.
Supported by the results here, we expect that our method can obtain a satisfactory performance even with incomplete ball trajectories, by using a simple interpolation as a work around to fill in the missing gaps.

\subsubsection{Visual Multi-modal Complementarity}

Since our approach is skeleton-based, it is important to assess how complementary it is with regard to other modalities, such as RGB and Optical Flow.
The experiments here target at demonstrating the complementarity capacity of our method.
For processing the visual information modalities (RGB and Flow) we use the Inception-v3~\cite{Szegedy2016} CNN architecture, following the Two-Stream approach~\cite{Simonyan2014}.

Each modality is trained separately and merged through a late fusion scheme, by combining the group activity predictions scores.
To classify the group activity using the CNNs, we separately input the cropped regions of all individuals, and use the output from the last layer before classification as features for each individual.
The individuals extracted features are then max-pooled and fed to a softmax layer for classification of the group activity. %
We train the RGB CNN using only the middle frame, and for test we pool the scores from the 10 frames in the center (middle frame, 5 frames before, and 4 after). 
For the Flow CNN, we train and test using a stack of Optical Flows extracted from the 10 central frames as describe above.
The RGB CNN model is initialized with pre-trained ImageNet weights, and the Flow CNN is trained from scratch.
When fusing the scores for the Visual modalities we set the weights to 2/3 and 1/3 for the RGB and Flow CNN respectively.
In the fusion of GIRN with RGB it is used equal weights for both (1/2), and in GIRN with Flow it is used 2/3 and 1/3.
For the GIRN and Visual fusion, we set the weights to 1/3 and 2/3 when fusing with $\text{GIRN}_{intra+inter}$, and equal weights for both (1/2) when fusing with $\text{GIRN}_{intra+inter+obj}$.
	
\begin{table}[ht]
	\centering
	\begin{threeparttable}
		\caption{Results for the experiments with distinct visual modalities and their fusion.}
		\begin{tabular}{lc}
			\toprule
			\textbf{Method}               & \textbf{Accuracy} \\ \midrule
			Appearance ($\text{CNN}_{RGB}$)                   & 88.5\% \\% InceptionV3 w/ pool 10
			Motion ($\text{CNN}_{Flow}$)                          & 79.7\% \\% InceptionV3 + new shuffle
			Visual ($\text{CNN}_{RGB}$ + $\text{CNN}_{Flow}$)         & 90.0\% \\% InceptionV3 .67 .33

			\midrule
			$\text{GIRN}_{intra+inter}$								 & 88.4\% \\ 
			$\text{GIRN}_{intra+inter}$ + Appearance			 & 91.3\% \\ %
			$\text{GIRN}_{intra+inter}$ + Motion				 & 89.8\% \\ %
			$\text{GIRN}_{intra+inter}$ + Visual				 & 93.0\% \\ %
			$\text{GIRN}_{intra+inter+obj}$							 & 92.2\% \\ 
			$\text{GIRN}_{intra+inter+obj}$ + Appearance		 & 93.5\% \\ %
			$\text{GIRN}_{intra+inter+obj}$ + Motion			 & 93.0\% \\ %
			$\text{GIRN}_{intra+inter+obj}$ + Visual			 & 94.0\% \\ %
			\bottomrule
		\end{tabular}
		
		\label{tab:volley_cnn}
	\end{threeparttable}
\end{table}

As it can be seen in Table~\ref{tab:volley_cnn}, the pose information from our approach shows high complementarity to the appearance and motion information coming from the CNN-based models.
Separately, our $\text{GIRN}_{intra+inter}$ model and the combined Visual CNN models ($CNN_{RGB}$ + $CNN_{Flow}$) have an accuracy of 88.4\% and 90.0\% respectively.
Fusing all three modalities leads to a significant improvement in performance, obtaining 93.0\%.
Moreover, our method that incorporates the person-object interactions can also be improved by fusing with the Visual CNNs, going from 92.2\% of accuracy to 94.0\%.  

\subsubsection{Comparison with Previous Work}

Here we compare previous techniques to our complete proposed method, using multiple relationships and including all components, the auxiliary individual modules and the attention mechanisms.
We chose to report for comparison the results from two of our multiple relations models, one using only the estimated poses and the other also including the annotated volleyball information, namely $\text{GIRN}_{intra+inter}$ and $\text{GIRN}_{intra+inter+obj}$ respectively.
The results for comparison are presented in Table~\ref{tab:volley_sota}.

\begin{table}[ht]
	\centering
	\begin{threeparttable}
	\caption{Comparison of our results with previous work on Volleyball dataset.}	
	\small
	\begin{tabular}{lccc} %
		\toprule
		\textbf{Method}               & \textbf{Accuracy}  & \normalsize{\textbf{Modalities}}  & \normalsize{\textbf{Backbone}} \\ \midrule
		HDTM~\cite{Ibrahim2016} & 81.9\% & RGB & AlexNet \\% CNN+ Ind-LSTM+ Grp-LSTM 
		CERN~\cite{Shu2017a} & 83.3\% & RGB & VGG-16 \\% CNN+ 3xLSTMs
		SSU~\cite{Bagautdinov2017} & 90.6\% & RGB & Inception-v3 \\% FCN (det. & feats) + RNN
		HRN~\cite{Ibrahim2018} & 89.5\% & RGB & VGG-19 \\% CNN + Hierarch. RelNets
		
		Multi-stream CNN~\cite{Azar2018} & 90.5\% & RGB+Flow+Pose$^\ast$ & Inception-v3 \\% Pose + RGB + OF
		
		PC-TDM~\cite{Yan2018} & 87.7\% & RGB + Flow & AlexNet \\% CNN + 3xLSTMs
		
		Multimodal Attention~\cite{Lu2019} & 91.7\% & RGB + Pose$^\dagger$ & Inception-v3 \\% ST-LSTM + CNN + GRU
		
		ARG~\cite{Wu2019a} & 92.6\% & RGB & VGG-16 \\% CNN + GCN
		CRM~\cite{Azar2019} & 93.0\% & RGB + Flow & I3D \\% Activity map, train ROI w/ openpose
		CCG-LSTM~\cite{Tang2019} & 89.3\% & RGB & AlexNet \\
		PMH~\cite{Chen2020} & 87.7\% & Pose$^\ast$ & ResNet-18 \\% Pose/Joint HeatMap + CNN
		
		Actor-Transformer~\cite{Gavrilyuk2020} & 94.4\% & RGB+Flow+Pose & I3D + HRNet \\ %
		HiGCIN~\cite{Yan2020a} & 91.4\% & RGB & ResNet-18 \\ 
		
		Context-Aware~\cite{Dasgupta2021} & 93.0\% & RGB+Pose & Inception-V3 + HRNet \\ %
		H-LSTM~\cite{Shu2021} & 88.4\% & RGB & AlexNet \\ 
		
		\midrule %
		$\text{GIRN}_{intra+inter}$ & 88.4\% & Pose & -- \\ 
		$\text{GIRN}_{intra+inter+obj}$ & \textit{92.2\%}
		& Pose & -- \\ 
		$\text{GIRN}_{intra+inter}$ + Visual & 93.0\% & RGB+Flow+Pose & Inception-v3 \\ %
		$\text{GIRN}_{intra+inter+obj}$ + Visual & \textit{94.0\%} & RGB+Flow+Pose & Inception-v3 \\ 
		\bottomrule
	\end{tabular}
	\begin{tablenotes}
		\scriptsize
        \item Pose$^\ast$ indicates the skeletons are transformed to a visual representation then fed to a CNN.
        \item Pose$^\dagger$ indicates the skeleton information is only indirectly used for description. %
	\end{tablenotes}
	
	\label{tab:volley_sota}
	\end{threeparttable}
\end{table}

The $\text{GIRN}_{intra+inter}$ form of our proposed method is already capable of outperforming many of the previous works (i.e. HDTM~\cite{Ibrahim2016}, CERN~\cite{Shu2017a}, PC-TDM~\cite{Yan2018} and PMH~\cite{Chen2020}), even though it does not incorporate any CNN-based input as they do (e.g. RGB, optical flow, heatmap).
Additionally, if we compare our method to reported ablation results not using RGB and OF, $\text{GIRN}_{intra+inter}$ also outperforms Multi-stream CNN~\cite{Azar2018} (82.6\%) and is more closely comparable to CRM~\cite{Azar2019} (90.8\%).

With regard to $\text{GIRN}_{intra+inter+obj}$, a direct comparison with previous work would be unfair because this variation makes use of manually annotated ball coordinates, data not automatically extracted from the videos and that does not contain errors.
It is a good indication, however, on the potential of such type of information, as can be seen that the explicit incorporation of this input by the relational modules is able to boost our method performance to an accuracy close to the best results obtained by the compared methods.
Moreover, this result is achieved while still keeping our input modality restricted to euclidean space representations, in other words, not incorporating visual spatial and motion information from RGB and optical flow.
In addition, our results related to robustness evaluation in Tables \ref{tab:pixel_noise} and \ref{tab:dropout_noise} shows our method can leverage the ball location information even if it is noisy or incomplete, and can still outperform many of the previous work under these conditions.

By fusing the GIRN with Visual information (RGB and Flow), our method can obtain even higher performance, more competitive with previous work results.
With 93.0\%, the $\text{GIRN}_{intra+inter}$ + Visual approach surpasses many previous works \cite{Wu2019a,Azar2018,Lu2019}.
Our $\text{GIRN}_{intra+inter+obj}$ + Visual method reaches up to 94.0\% and is comparable to Actor-Transformer~\cite{Gavrilyuk2020}.
We would like to highlight that our approach shows good complementarity to the visual information even though the visual component employed by us here is relatively simple and straightforward.
We just fine-tune a traditional 2D CNN architecture on top of the RGB and Flow data.
Different from previous works, which employ more advanced techniques such as LSTMs, GCNs, 3D CNNs, and their own proposed methods. 

Although not actively targeting at it, it is possible that the CNN image-based approaches are already learning patterns related to the ball location and movement during the activity.
Therefore, this source of information should be somehow explicitly incorporated to pose-based approaches, for them to be able to suitably reason about group activity on sports.
Besides, even the non-pose approaches could possibly improve their performance if the ball information was deliberately explored in their architecture.
Naturally this should also hold true to different scenarios other than sports, whenever there are relevant objects that the individuals are directly or indirectly interacting with. %

\subsubsection{Qualitative Analysis}

\begin{figure}[!ht]
	\centering
	\makebox[\textwidth][c]{\includegraphics[width=1.05\textwidth,trim={10 0 10 0},clip]{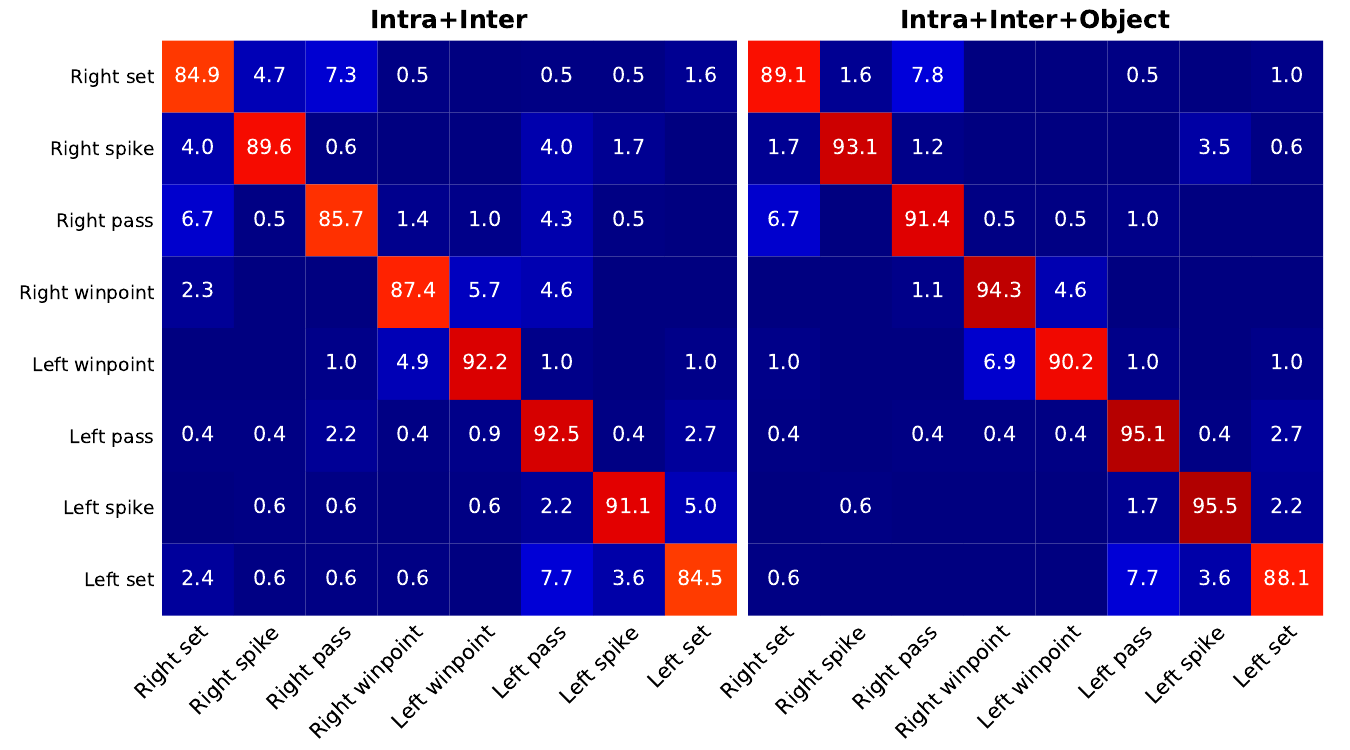}}%
	\caption{Confusion matrices for our methods variations using pose-only ($\text{GIRN}_{intra+inter}$) and pose plus volleyball ($\text{GIRN}_{intra+inter+obj}$).}
	\label{fig:confusion_matrices}
\end{figure}

For a more in depth analysis of the performance of our method, and also to visualize where are the differences in the results from our variations with and without using the volleyball, we produce the confusion matrices in Fig.~\ref{fig:confusion_matrices}. 

Focusing first in the confusion matrix for $\text{GIRN}_{intra+inter}$, it can be seen a consistent confusion between Right and Left sides for the same type of activities (anti-diagonal values). The degree of confusion changes slightly for some of the activities (higher with Pass) and the sides (higher for Right).
Moving on to $\text{GIRN}_{intra+inter+obj}$ matrix, it is evident that the addition of the ball information greatly helps at distinguishing between sides. 
There are much less confusion in the anti-diagonal cells and in its containing quadrants, only a few exceptions remains and with low-values mostly.
However for the Winpoint activity, the confusion is still high between sides.
Such result is not surprising, since for this activity the side label is not straightforwardly related to the location of the ball, in many cases the ball is not even present in the scene.
For this activity there are instances which are challenging even for humans to distinguish, requiring a longer term inference based on which team scored the point, or a finer analysis per individual to identify if their response is cheerful or upset.

With regard to the activities directly involving the volleyball, Spike seems to be more distinguishable, obtaining the highest accuracy results in average. 
Pass and Set in the other hand have a greater confusion between them, and also some miss-labeling for Set as Spike.
A possible cause for these confusions are that the pose estimation might have some errors -- due to model precision or occlusion -- in the hands and arms joints, which are more important for identifying these activities.
Such issues might be automatically sorted out when more precise pose estimation models are available. %
Another reason might be temporally related. 
These activities often happen in a sequence (i.e. Pass $\Rightarrow$ Set $\Rightarrow$ Spike), and the temporal length of our input comprises the whole snippet duration, thus it is possible our method is labeling the scene with the previous or next activity instead of the central one.
This could be solved by adding a mechanism to learn the temporal dynamics and causality between actions, or by giving more weight to the central activity. %

\section{Conclusion}
\label{sec:conclusion}

In this paper, we tackled the problem of group activity recognition by a different perspective than previous works: using skeleton information for reasoning over the interactions between multiple individuals and also between individuals and objects.
We demonstrated, through our GIRN, that such pose information can be leveraged for solving this problem by using multiple pair-wise relational modules that will specialize in different types of relationships based on the nature of the pairs being fed to them.
With the support of individual actions knowledge to refine the relations' descriptions, and attention mechanisms to give more importance to key persons, our method obtain promising results.
Although being exclusively based in a single modality for input, our method is able to achieve competitive performance with respect to the state-of-the-art, which commonly uses at least RGB and Optical Flow modalities.
Our experiments also demonstrate how explicitly accommodating domain specific objects of interest (e.g. volleyball) in the architecture can lead to significant improvements to performance. 
To the best of our knowledge, this is the first work to actively explore such type of information for the task of group activity recognition.

Currently, the GIRN is operating at the complete scene temporal range as a single input, this is possible because the activities for the volleyball dataset are defined within short duration clips. 
However, the performance might be hindered if handling longer duration activities. To overcome this, temporal dynamics learning should be incorporated to the architecture, for example through GRUs or LSTMs. The use of such temporal aware techniques could even improve the performance for the shorter cases also, given that they may help finding some causality between individual actions, and by making the model more robust to temporal displacements.

Our proposed method can also be improved through other manners.
For example by feeding the relations descriptions extracted by our relational modules to a more complex higher-level inference representation, such as graphs, this way allowing the use of graph-based techniques also.
Another manner for improvement of our method would be to fuse the visual information at an earlier stage, to improve the relationship learning and the individuals description before the group activity inference module.
An additional way to improve our method is by exploring other objects of interest, for example the volleyball net, or even landmark spatial locations, such as the borders of the volleyball court.

\section*{Acknowledgment}

This research was carried out at the Rapid-Rich Object Search (ROSE) Lab, Nanyang Technological University (NTU), Singapore and supported by a grant from NTU’s College of Engineering (M4081746.D90).

\bibliography{references}

\end{document}